\documentclass[12pt,a4paper]{article}

\usepackage[colorlinks,urlcolor=blue,linkcolor=blue,citecolor=blue]{hyperref}
\expandafter\def\expandafter\UrlBreaks\expandafter{\UrlBreaks\do\/\do\*\do\-\do\~\do\'\do\"\do\-}
\usepackage{color}
\usepackage[top=1.2in, left=1in, right=1in, bottom=1.2in]{geometry}

\usepackage{amsmath,amsfonts}
\usepackage{algorithmic}
\usepackage{array}
\usepackage[caption=false,font=normalsize,labelfont=sf,textfont=sf]{subfig}
\usepackage{textcomp}
\usepackage{stfloats}
\usepackage{url}
\usepackage{verbatim}
\usepackage{graphicx}
\def\BibTeX{{\rm B\kern-.05em{\sc i\kern-.025em b}\kern-.08em
    T\kern-.1667em\lower.7ex\hbox{E}\kern-.125emX}}
\usepackage{balance}
\usepackage{multirow}
\usepackage{booktabs}
\usepackage{makecell}
\usepackage{lscape}

\usepackage{tgtermes}

\newenvironment{Biography}[1]{\noindent\textbf{#1} \,}{\vspace{2mm}}

\newcommand{\notable}[1]{#1}

\newcommand{\resu}[1]{\underline{#1}}
\newcommand{\resb}[1]{\textbf{#1}}
\newcommand{\resub}[1]{\underline{\textbf{#1}}}

\newcommand{\doi}[1]{\href{https://doi.org/#1}{#1}}
\newcommand{\arxiv}[1]{arXiv: \href{#1}{#1}}

\newcommand{\percent}{\%}

\begin{document}

\title{\textbf{\fontsize{20pt}{22pt}\selectfont MangaUB: A Manga Understanding Benchmark \\for Large Multimodal Models}}
\date{}

\author{Hikaru Ikuta,
Leslie Wöhler, and
Kiyoharu Aizawa\\
{\small The University of Tokyo, 113-8656, Tokyo, Japan}}

\maketitle

\begin{abstract}
\looseness-1
Manga is a popular medium that combines stylized drawings and text to convey stories. As manga panels differ from natural images, computational systems traditionally had to be designed specifically for manga. Recently, the adaptive nature of modern large multimodal models (LMMs) shows possibilities for more general approaches. To provide an analysis of the current capability of LMMs for manga understanding tasks and identifying areas for their improvement, we design and evaluate MangaUB, a novel manga understanding benchmark for LMMs. MangaUB is designed to assess the recognition and understanding of content shown in a single panel as well as conveyed across multiple panels, allowing for a fine-grained analysis of a model's various capabilities required for manga understanding. Our results show strong performance on the recognition of image content, while understanding the emotion and information conveyed across multiple panels is still challenging, highlighting future work towards LMMs for manga understanding.
\end{abstract}

\section{Introduction}

Manga is an art form that conveys stories using a mixture of text and images.
Its textual and visual language differs not only from natural images but also from American comics~\cite{pratha2016pow}.
Works of manga reach a worldwide audience with simultaneous release in many countries,
highlighting their popularity and economic significance.
There is thus a high demand for computational systems that support the creation and publication process, for example enabling automatic translation~\cite{hinami2021towards}.

Recently, large multimodal models (LMMs) have demonstrated outstanding performance on various image understanding tasks.
A significant application of LMMs is their role as foundational models, where they are integrated as modules to process inputs for downstream tasks.
We anticipate that advanced models for manga understanding tasks will adopt this approach.
Consequently, it is crucial to design a system that can thoroughly evaluate an LMM's manga understanding capabilities for future research.

\notable{
\begin{figure*}[!th]
    \centering
    \includegraphics[width=6.3in]{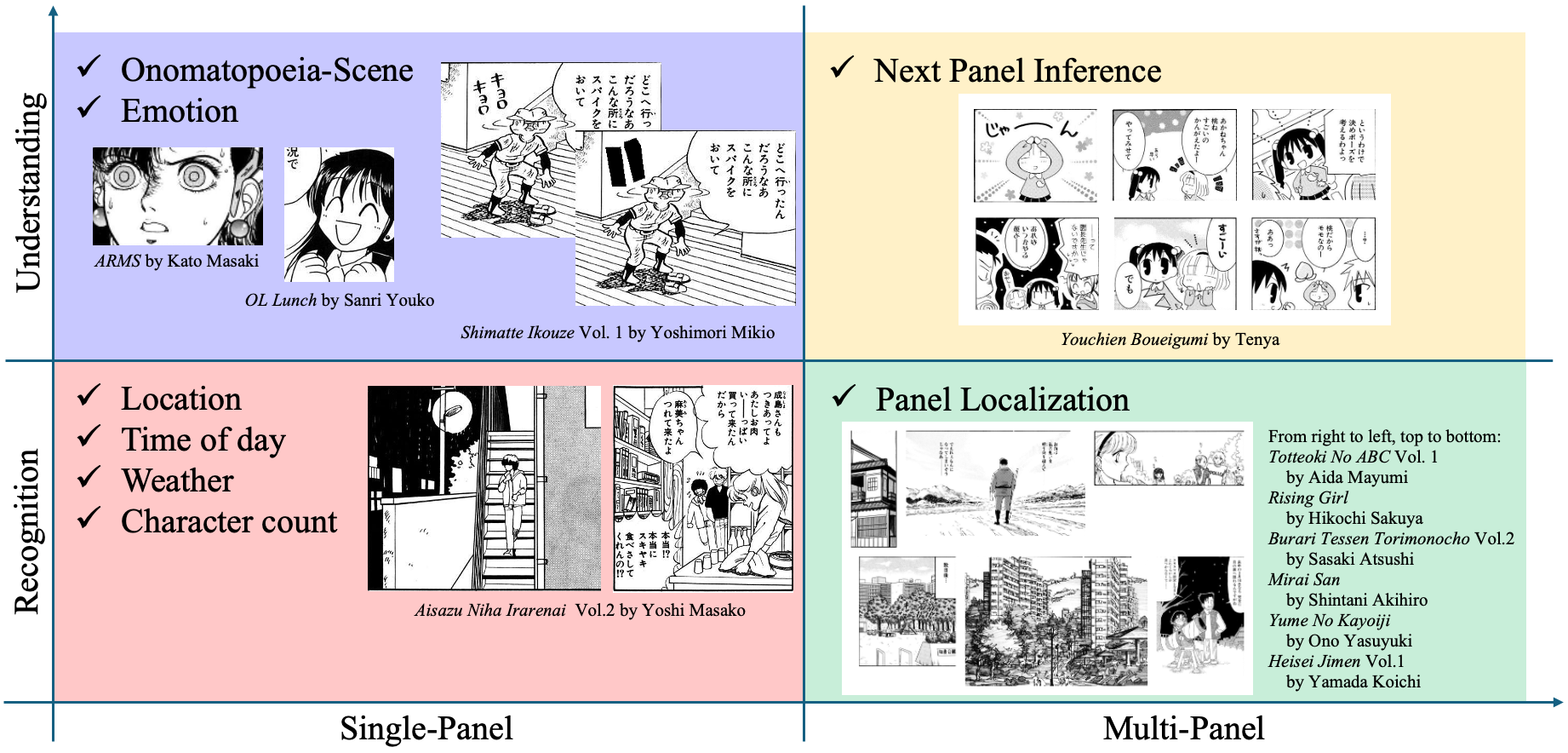}
    \caption{A summary of the tasks in the benchmark. Our tasks focus on recognition and understanding of image content for single-panel and multi-panel input.
    All manga panel images in this figure are referenced from the Manga109 dataset~\cite{mtap_matsui_2017}, courtesy of the authors of each manga indicated in the figure.}
    \label{fig:quadrants}
\end{figure*}

\begin{figure*}[!th]
    \centering
    \includegraphics[width=5.8in]{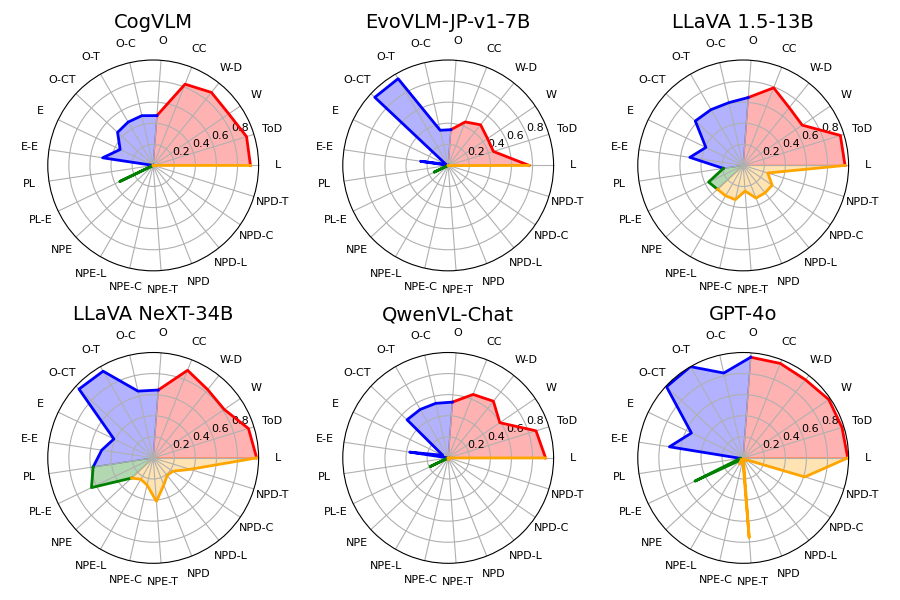}
    \caption{Performances for various open-source and proprietary models.
    Results of the evaluation of various open-source and proprietary models indicating strong performance on the recognition of image content, while understanding the emotion and information conveyed across multiple panels is challenging.
    The abbreviations used in the labels are described in Table~\ref{tab:prompt-count-stats}.}
    \label{fig:radarchart}
\end{figure*}
}

In order to understand how LMMs could be used to improve manga-related computation systems,
we propose MangaUB, a manga scene understanding benchmark for LMMs.
Reflecting the fact that understanding manga requires various layers of cognitive tasks,
our benchmark tasks are divided into four quadrants, as shown in Figure~\ref{fig:quadrants}.
The benchmark includes various tasks for image recognition and understanding, as well as processing single panels and multiple panels, allowing for a fine-grained analysis of the various capabilities required for manga understanding.
The benchmark contains 6,585 distinct questions with 18,179 prompts, with additional prompts included for taking prompt sensitivity into account.
For constructing the dataset, 2,967 human-provided annotations were created for this benchmark.
The annotations created for this benchmark along with the code will be made publicly available.

The contributions in this paper are the following:
\begin{itemize}
    \item Proposal of MangaUB, a manga scene understanding benchmark for LMMs, which will be made publicly available.
    \item Collection of a manga dataset representing four demographics and providing 2,967 new human annotations of domain experts.
    \item Evaluation and comparison of various LMMs identifying current shortcomings and directions for future work.
\end{itemize}

\section{Related Work}

\subsection{Computational Analysis of Manga}
The style of manga possesses a distinct visual language referring to all aspects of the medium
such as sound effects~\cite{pratha2016pow} and page layout~\cite{cohn2017cultural}.
For computational systems, the differences between manga and natural images traditionally made it necessary to create specialized systems for their recognition and analysis~\cite{augereau2018survey}, with techniques focusing on the detection of panels and extraction of text~\cite{sharma2024image}, or to provide reasoning about panel content~\cite{Iyyer2017CVPR}.
Furthermore, datasets and benchmarks consisting of manga have been introduced to coordinate research and adhere to copyright standards~\cite{multimedia_aizawa_2020}. Some datasets focus on the visual language of manga like understanding sound effects~\cite{comiconomdataset2022} and characters' emotion~\cite{KangaiSet}.

Previous manga-related benchmarks evaluate the recognition of written texts~\cite{sachdeva2024magi} and the performance of visual closure~\cite{Iyyer2017CVPR}.
A concurrent work~\cite{vivoli2024comicsdatasetsframeworkmix} present a compilation of manga benchmarks, but their focus is on detection.
As our benchmark focuses on identifying the narrative conveyed by the combinations of the content depicted inside a panel or a series of panels,
our focus extends beyond the previous tasks of identifying the individual elements within the image such as the text, characters, and speech bubble association discussed in ~\cite{sachdeva2024magi}.
The present paper also focuses on the performance of LMMs, which has not been discussed in these previous works.
The present benchmark allows for a fine-grained analysis of the model's capabilities,
providing various insights for the future rooms of improvement for LMMs.

\subsection{LMM Benchmarks}
One of the difficulties in evaluating the performance of LMMs is managing its broad output space.
Existing benchmarks handle this problem mainly in two ways:
(1) imposing a constraint in the expected output space, e.g., by using multiple-choice questions~\cite{liu2024mmbenchmultimodalmodelallaround}, or
(2) using an additional language model to compare the model's output with the reference answer.
In our benchmark the majority of the tasks take the format of multiple-choice questions.

\section{Benchmark Design Overview}
The process of reading manga involves multiple cognitive tasks.
Reflecting this, the benchmark tasks are divided along two dimensions into four quadrants, as shown in Figure~\ref{fig:quadrants}.
Scene Understanding tasks require a higher level of scene comprehension compared to Scene Recognition tasks.
Single-Panel tasks involve one panel or a cropped part of a panel, while Multi-Panel tasks require comprehension and comparison of multiple panels.
This design allows for a fine-grained analysis of the manga comprehension abilities of LMMs.

\notable{
\begin{table*}[!t]
    \fontsize{8.7pt}{10.7pt}\selectfont
    \caption{The question and prompt count statistics of the entire benchmark.
    \#Questions indicates distinct questions.
    \#Prompts indicates the total number of prompts used.
    Multiple-choice question tasks use multiple prompts for each question to account for prompt sensitivity,
    using the CircularEval scheme.
    In the Emotion and Panel Localization tasks, different abbreviations are used for the metric used for the score.
    E and PL indicate CircularEval,
    while E-E and PL-E indicate Ensemble Accuracy.
        \label{tab:prompt-count-stats}}
    \vspace{8.7pt}
    \centering
    \begin{tabular}{llllc|ll}
    \toprule
Group                      & Task                    & Category      & Abbreviation & \#Choices  & \#Questions   & \#Prompts       \\\midrule
Single-Panel Recognition   & Location                &               & L            & 2          & 953           & 1906            \\
                           & Time of Day             &               & ToD          & 2          & 512           & 1024            \\
                           & Weather                 & Normal        & W            & 2          & 177           & 354             \\
                           &                         & Difficult     & W-D          & 3          & 197           & 591             \\
                           & Character Count         &               & CC           & $n \geq 0$ & 1128          & 1128            \\\midrule
Single-Panel Understanding & Onomatopoeia Scene      & Baseline      & O            & 3          & 101           & 303             \\
                           &                         & Cropped       & O-C          & 3          & 101           & 303             \\
                           &                         & With Text     & O-T          & 3          & 101           & 303             \\
                           &                         & Crop+Text     & O-CT         & 3          & 101           & 303             \\
                           & Emotion                 & Face          & E/E-E        & 7          & 168           & 1176            \\
                           &                         & Body          &              & 7          & 168           & 1176            \\\midrule
Multi-Panel Recognition    & Panel Localization      & Location      & PL/PL-E      & 6          & 190           & 1140            \\
                           &                         & Time-of-day   &              & 6          & 101           & 606             \\
                           &                         & Weather       &              & 6          & 35            & 210             \\\midrule
Multi-Panel Understanding  & Next Panel Easy         & Baseline      & NPE          & 3          & 319           & 957             \\
                           &                         & Left-First    & NPE-L        & 3          & 319           & 957             \\
                           &                         & Cropped       & NPE-C        & 3          & 319           & 957             \\
                           &                         & With Text     & NPE-T        & 3          & 319           & 957             \\
                           & Next Panel Difficult    & Baseline      & NPD          & 3          & 319           & 957             \\
                           &                         & Left-First    & NPD-L        & 3          & 319           & 957             \\
                           &                         & Cropped       & NPD-C        & 3          & 319           & 957             \\
                           &                         & With Text     & NPD-T        & 3          & 319           & 957             \\\midrule
Total                      &                         &               &              &            & 6585          & 18179           \\
\bottomrule
\end{tabular}
\end{table*}

\begin{table*}[!t]
    \fontsize{10pt}{11pt}\selectfont
\caption{Per-label data count statistics of the Single-Panel tasks.
All of the panel images and panel bounding box annotations, including the Onomatopoeia and Emotion tasks, are from the Manga109 dataset~\cite{mtap_matsui_2017}.
``New'' in ``Annotation Source'' indicates labels that are newly provided in this paper.
\label{tab:task-samples-table}}
\vspace{10pt}
\centering
\begin{tabular}{lll|l|llll}
    \toprule
Task               & Annotation Source                                                   & Label     & Total  & Shonen   & Shojo   & Seinen   & Josei \\\midrule
Location           & New                                                                 & Indoors   & 442    & 111      & 110     & 107      & 114   \\
                   &                                                                     & Outdoors  & 511    & 166      & 103     & 131      & 111   \\\midrule
Time of Day        & New                                                                 & Day       & 228    & 69       & 54      & 53       & 52    \\
                   &                                                                     & Night     & 284    & 79       & 65      & 84       & 56    \\\midrule
Weather            & New                                                                 & Sunny     & 174    & 66       & 32      & 46       & 30    \\
                   &                                                                     & Rainy     & 180    & 30       & 44      & 76       & 30    \\
Weather-Difficult  &                                                                     & Snowy     & 20     & 5        & 4       & 11       & 0     \\\midrule
Character Count    & New                                                                 & 0         & 146    & 28       & 46      & 30       & 42    \\
                   &                                                                     & 1         & 303    & 82       & 71      & 73       & 77    \\
                   &                                                                     & 2         & 305    & 78       & 76      & 77       & 74    \\
                   &                                                                     & 3         & 237    & 67       & 53      & 55       & 62    \\
                   &                                                                     & 4         & 137    & 39       & 29      & 35       & 34    \\\midrule
Subtotal           &                                                                     &           & 2967   & 820      & 687     & 778      & 682   \\\midrule
Onomatopoeia       & COO~\cite{comiconomdataset2022} + Table~\ref{tab:onom-defs} (New)   &           & 101    & 58       & 16      & 19       & 8     \\\midrule
Emotion            & KangaiSet~\cite{KangaiSet}                                          & Anger     & 24     & 6        & 6       & 6        & 6     \\
                   &                                                                     & Disgust   & 24     & 6        & 6       & 6        & 6     \\
                   &                                                                     & Fear      & 24     & 6        & 6       & 6        & 6     \\
                   &                                                                     & Happiness & 24     & 6        & 6       & 6        & 6     \\
                   &                                                                     & Neutral   & 24     & 6        & 6       & 6        & 6     \\
                   &                                                                     & Sadness   & 24     & 6        & 6       & 6        & 6     \\
                   &                                                                     & Surprise  & 24     & 6        & 6       & 6        & 6     \\\midrule
Total              &                                                                     &           & 3236   & 920      & 745     & 839      & 732   \\
    \bottomrule
\end{tabular}
\end{table*}
}

\subsection{The Input Design of the Multi-Panel Tasks}
The spatial arrangement of manga panels is used as a part of the visual language of manga~\cite{cohn2017cultural}.
Therefore, understanding the two-dimensional layouts of panels is an important capability in advanced manga understanding systems.
We have thus chosen the design of the input images in the Multi-Panel tasks so that multiple panel are shown to the model as one image (see Figure~\ref{fig:quadrants}).
Similar input methods are used in various previous works on LMMs, such as work on user interface recognition~\cite{baechler2024screenaivisionlanguagemodelui}.

\subsection{Evaluation Methods}
Following previous work on LMM benchmarks~\cite{liu2024mmbenchmultimodalmodelallaround},
multiple-choice tasks in this benchmark are evaluated using the CircularEval scheme.
This scheme considers prompt sensitivity by evaluating each question with multiple prompts,
rewarding consistent responses.
For each multiple-choice question with $n$ choices, $n$ prompts are created,
cyclically permutating the choices so the correct answer appears in all $n$ indices.
The model must answer all prompts correctly for the question to be counted as correct.
The final score is the ratio of questions marked as correct to the total number of questions, similar to the accuracy metric.

In the MMBench benchmark~\cite{liu2024mmbenchmultimodalmodelallaround} where CircularEval was proposed,
the number of choices for each question ranged from two to four.
In our work, some of the tasks contain up to seven choices,
increasing the strictness of CircularEval.
To address this, we introduce the Ensemble Accuracy metric for questions with more than four choices,
allowing for a more detailed analysis.
This metric uses the same set of questions as CircularEval but takes the majority answer as the model’s final output.
A question is marked as correct if the majority is unique and correct.
This additional metric provides a more detailed analysis for challenging tasks with many choices.

\subsection{Models}
We experiment on open-source LMMs and one proprietary LMM.
For the open-source LMMs, we choose ones that are released recently:
CogVLM-17B~\cite{wang2024cogvlmvisualexpertpretrained} EvoVLM-JP-v1-7B~\cite{akiba2024evolutionaryoptimizationmodelmerging}
Qwen-VL-Chat~\cite{bai2023qwenvlversatilevisionlanguagemodel}
LLaVA-1.5-13B~\cite{Liu_2024_CVPR}
LLaVA-NeXT-34B~\cite{liu2024llavanext}.
For proprietary LMMs, we choose GPT-4o (gpt-4o-2024-05-13) published by OpenAI.

EvoVLM-JP-v1-7B is a model based on the Evolutionary Model Merge method~\cite{akiba2024evolutionaryoptimizationmodelmerging},
trained in a way to be capable of math reasoning and Japanese LMM tasks.
Note that all prompts are written in English, except for a limited number of tasks which contain text annotations in Japanese, which will be described in detail in the next section.

In order to facilitate a consistent evaluation, we disable sampling for all open-source models to reduce the randomness in the output.
For GPT-4o, we set the temperature hyperparameter to 0.
All experiments for the open-source models were performed on NVIDIA A100 40GB GPUs.

\section{Task Definition Details}
Here we will describe each task in detail, according to the four quadrants shown in Figure~\ref{fig:quadrants}.
Table~\ref{tab:prompt-count-stats} summarizes the total number of distinct questions and prompts in each task.

\subsection{Single-Panel Scene Recognition Tasks}
The scene recognition tasks are designed to capture the model's ability in recognizing the basic elements of the scene and panel layout.

The panel bounding box labels and manga images are referenced from the Manga109 dataset~\cite{mtap_matsui_2017,multimedia_aizawa_2020}.
All of the labels for each panel in the tasks in this category were newly annotated by an expert in our team.
There are 2,967 newly provided annotations created for this benchmark.
As manga can differ in style based on the targeted demographic, we include panels from various genres, namely Shonen, Shojo, Seinen, and Josei manga.
Detailed data counts for the annotations are summarized in Table~\ref{tab:task-samples-table}.
All of these annotations will be made publicly available.

\noindent\textbf{Location.} Identify whether the location depicted in the manga panel is indoors or outdoors.

\noindent\textbf{Time of Day.} Identify whether the time of day in the scene depicted in the manga panel is day or night.

\noindent\textbf{Weather.} Identify the weather of the scene. There are two categories, Normal and Difficult, based on the number of choices. The Normal category has two choices ``sunny'' and ``rainy,'' and the Difficult category has three choices, ``sunny,'' ``rainy,'' and ``snowy.''
The separation is due to the scarce instances of ``snowy.''

\noindent\textbf{Character Count.} Identify the number of characters (people) shown in the panel. The model is expected to provide the answer as one number denoted in Arabic numerals.
The ground truth labels are between 0 and 4 inclusive, but the prompt does not explicitly state these bounds.

This is the only task that is not designed as a multiple-choice question in this benchmark.
Since CircularEval does not apply for this task,
we measure the accuracy as the final benchmark score.

\subsection{Single-Panel Scene Understanding Tasks}
For scene understanding, we design tasks that require not only the recognition of the scene elements but also the comprehension of elements that help in understanding the narrative flow of scenes and the author's intention.

\notable{
\begin{table*}[!t]
  \fontsize{10pt}{12pt}\selectfont
  \caption{Onomatopoeia descriptions used in the prompts.
  Descriptions were written based on a Japanese-English onomatopoeia dictionary~\cite{onomdictionary}.
  \label{tab:onom-defs}}
  \vspace{10pt}
  \centering
  \begin{tabular}{ll|l}
  \toprule
  Onomatopoeia & \#Data  &  Scene Description \\
  \midrule
  Zawa         & 17      &  The noise of a crowd fills the scene. \\
  Waa          & 10      &  A character or a crowd is shouting. \\
  Za           & 10      &  A sudden sound is made. \\
  Kyaa         & 8       &  A character or a crowd is screaming. \\
  Shin         & 8       &  The scene is silent. \\
  Kusu         & 8       &  A character or a crowd is giggling, chuckling, or snickering. \\
  Kyoro        & 7       &  A character is looking around curiously. \\
  Ha           & 7       &  A character is startled, and their breath is taken away. \\
  Mogu         & 7       &  A character is munching on some food. \\
  Gacha        & 7       &  A clattering or rattling sound is made. \\
  Niko         & 6       &  A character is smiling and looking happy. \\
  Biku         & 6       &  A character is startled or alarmed. \\
  \bottomrule
\end{tabular}
\end{table*}
}

\noindent\textbf{Onomatopoeia Scene.}
This task measures how much the model can understand the relevant context of a given scene.
The task is formulated as a three-choice problem where the model chooses the most fitting scene description for a given manga panel.

Designing manga scene description tasks is challenging due to the need for relevant story descriptions.
To avoid bias towards particular LMMs, human-generated labels are ideal but time-consuming.
To address this, our benchmark employs a novel pipeline using
\textit{onomatopoeia}, or the sound effects drawn in the manga panels,
which are one of the key elements in conveying actions and the mood of a scene~\cite{pratha2016pow}.
Not only can onomatopoeia highlight audible sound effects, but can also show actions, and even the absence of sound,
demonstrating their diverse information (see Table~\ref{tab:onom-defs}).
Most importantly, as onomatopoeia are deliberately added by the authors,
it could strongly be expected that they convey crucial narrative information,
forming the basis of our scene description generation method.

We use the annotations from the Comic Onomatopoeia Dataset (COO Dataset)~\cite{comiconomdataset2022}.
We first hand-select a pool of 12 onomatopoeia shown in Table~\ref{tab:onom-defs},
all within the top 3~\percent in the occurrence ranking of the COO Dataset.
Occurrences are measured by exact text match, treating the Hiragana and Katakana notational differences of the Japanese language as separate labels.
We then select 101 panels containing the required onomatopoeia.
Descriptions were written by an expert in our group, based on a Japanese-English onomatopoeia dictionary~\cite{onomdictionary}.
The sentences used in the prompts and the number of data points are summarized in Table~\ref{tab:onom-defs}.

The task includes four conditions for an ablation study:
(1) Baseline: Shows the manga panel as is, without any editing.
(2) Cropped: All onomatopoeia are cropped out using polygon region annotations from the COO Dataset.
(3) With Text: Transcriptions of all onomatopoeia in the panel are added to the prompt, written in Japanese and quoted in Japanese quotation marks. The rest of the prompt is in English.
(4) Crop+Text: Combines conditions 2 and 3.

The final benchmark score is the the macro average of the CircularEval metric for each onomatopoeia.

\noindent\textbf{Emotion.}
This task aims to capture the model's ability in understanding the emotions conveyed by a character's expression.
The data is gathered from the KangaiSet dataset~\cite{KangaiSet}.
In this task, the model must select the emotion of a given character from seven choices: Anger, Disgust, Fear, Happiness, Sadness, Surprise, and Neutral.
An example input image is shown in the top left corner of Figure~\ref{fig:quadrants}.

The original KangaiSet dataset contains 9,387 labels based on human judgments.
However, the data is unbalanced, with only 44 samples for ``disgust.''
To balance the data, two experts from our group selected 168 images from KangaiSet,
evenly split among the seven emotions and four genres as shown in Table~\ref{tab:task-samples-table}.

Based on the design from the KangaiSet dataset, we set two categories for this task:
(1) Face: Only the character's face is shown.
(2) Body: The visible area extends to the character's body, using ``body'' bounding box annotations from the Manga109 dataset.
We excluded the ``Panel'' category from KangaiSet due to ambiguity when multiple characters appear in one panel.

We use the CircularEval and Ensemble Accuracy metrics for this task.
Metrics are calculated for the Face and Body categories, and the final benchmark score is the macro average of these metrics.

\noindent\subsection{Multi-Panel Recognition Task}

\noindent\textbf{Panel Localization.}
This task measures the model's ability to understand the two-dimensional layout of panels.

We arrange six manga panels into two rows of three panels, as shown in the bottom right corner of Figure~\ref{fig:quadrants}.
Panels and labels are sourced from the Location, Time of Day, and Weather tasks, yielding a total of six different labels.
Among the six panels, five share the same label, while one has the opposite label.
The model must identify the panel with a specified label.
The negative and positive labels always belong to the same task.

Panels are specified as
``Top left,''
``Top middle,''
``Top right,''
``Bottom left,''
``Bottom middle,''
and ``Bottom right'' in the choices.
The order of the choices is randomized for each question and permutated six times to measure the CircularEval metric.

To balance the influence of the three base Single-Panel tasks, we group the questions according to which label is used as the query and which position is used as the answer.

We use the CircularEval and Ensemble Accuracy metrics for this task.
Metrics are calculated for each group, and the final benchmark score is the group macro average of each metric.

\subsection{Multi-Panel Understanding Task}

\noindent\textbf{Next Panel Inference.}
This tasks aims to measure the model's ability in understanding the connection between panels and the underlying narrative structure based on the \textit{Visual Cloze} task by Iyyer et al.~\cite{Iyyer2017CVPR}.

In this task, three consecutive panels are presented as the context, and the model must choose the panel that immediately follows after the given three context panels.
Both the context and choice panels are shown in one image,
as shown in the top right corner of Figure~\ref{fig:quadrants}.
Following the reading order of manga, the panels are read from right to left, which is specified in the prompt.
The choices are ``Bottom left,'' ``Bottom middle,'' and ``Bottom right,'' with the order of the choices randomized for each question.
This two-by-three panel arrangement aligns with the Panel Localization task, enabling separate analysis of spatial recognition and story comprehension.
The final benchmark score is the macro average of the CircularEval metrics for each choice.

As selecting panel sequences at random as in Iyyer et al. could lead to sequences that are only loosely connected, we use four-panel manga for this task.
In a typical four-panel manga, one short story is told in exactly four panels that are usually self-contained,
aligning with the purpose of the task.
We selected all five volumes from the Manga109 dataset~\cite{mtap_matsui_2017,multimedia_aizawa_2020} labeled as ``four-frame cartoons''
and manually curated four-panel sequences, removing any that did not consist of exactly four panels.

To allow for future training, we split the dataset into train, validation, and test sets, with 367, 47, and 319 four-panel manga sequences, respectively.
The test set is used for the benchmark.
Four volumes are split 7:3 into (train + validation) : test, and one volume is included entirely in the test set.
This split allows for the assessment of performance on both known and unknown manga when the training dataset is used.

We prepare two difficulties, Easy and Difficult, similar to Iyyer et al:
(1) Easy: The negative examples are randomly sampled from different manga volumes,
(2) Difficult: The negative examples are randomly sampled from from the same manga volume.

For each difficulty, we set four different conditions to allow for an ablation study:
(1) Baseline: The base task as described previously.
(2) Left First: The panels in the input image is read from left to right.
(3) Cropped: The text in all context and choice panels are cropped out and filled in black.
(4) Crop+Text: Along with cropping, the contents of the cropped out text are included in the prompt text.

In Crop+Text, the bounding box and text annotations from Manga109 are used. The text is all written in Japanese. Each text annotation is quoted in Japanese quotation marks. Note that the rest of the prompt is written in English.
For each panel, the texts are ordered according to the $x$ coordinate of the rightmost corner of their bounding boxes,
reflecting the right-to-left, top-to-bottom text reading order of Japanese manga.

\section{Results and Discussions}

\notable{
\begin{table*}[!t]
    \fontsize{8pt}{9.5pt}\selectfont
    \caption{An overview of the benchmark results.
    Boldface numbers indicate the highest score among all models.
    Underlined numbers indicate the highest score among the open-source models.
    In the ``Metric'' column, ``Circ. Acc.'' indicates the CircularEval metric,
    and ``Ens. Acc.'' indicates the Ensemble Accuracy metric introduced in the ``Evaluation Methods'' section.\label{tab:benchmark-overview}}
    \vspace{8pt}
    \centering
    \begin{tabular}{lccc|ccccc|c}
    \toprule
Task                    & Category      & Metric              & Choices    & CogVLM            & \makecell{EvoVLM\\JP-v1-7B}   & \makecell{LLaVA\\1.5-13B}   & \makecell{LLaVA\\NeXT-34B}    & \makecell{Qwen-VL\\Chat} & GPT-4o           \\\midrule
Location                &               & Circ. Acc.          & 2          & 0.923             & 0.768                         & 0.966                       & \resu{0.982}                  & 0.924                    & \resb{0.992}     \\
Time of Day             &               & Circ. Acc.          & 2          & 0.927             & 0.449                         & \resu{0.965}                & 0.946                         & 0.872                    & \resb{0.981}     \\
Weather                 & Normal        & Circ. Acc.          & 2          & \resu{0.868}      & 0.452                         & 0.678                       & 0.816                         & 0.593                    & \resb{0.983}     \\
                        & Difficult     & Circ. Acc.          & 3          & \resu{0.886}      & 0.493                         & 0.701                       & 0.829                         & 0.688                    & \resb{0.948}     \\
Character Count         &               & Acc.                & $n \geq 0$ & 0.828             & 0.443                         & 0.791                       & \resu{0.893}                  & 0.647                    & \resb{0.963}     \\\midrule
Onom. Scene             & Baseline      & Circ. Acc.          & 3          & 0.474             & 0.340                         & \resu{0.646}                & \resu{0.646}                  & 0.531                    & \resb{0.958}     \\
                        & Cropped       & Circ. Acc.          & 3          & 0.484             & 0.342                         & 0.611                       & \resu{0.650}                  & 0.531                    & \resb{0.826}     \\
                        & With Text     & Circ. Acc.          & 3          & 0.475             & \resu{0.950}                  & 0.611                       & \resu{0.950}                  & 0.531                    & \resb{1.000}     \\
                        & Crop+Text     & Circ. Acc.          & 3          & 0.461             & 0.950                         & 0.620                       & \resu{0.960}                  & 0.532                    & \resb{0.992}     \\
Emotion                 &               & Circ. Acc.          & 7          & 0.348             & 0.024                         & 0.393                       & \resu{0.414}                  & 0.057                    & \resb{0.545}     \\
                        &               & Ens. Acc.           & 7          & 0.485             & 0.265                         & \resu{0.512}                & 0.494                         & 0.369                    & \resb{0.708}     \\\midrule
Panel Loc.              &               & Circ. Acc.          & 6          & 0.000             & 0.000                         & 0.190                       & \resub{0.576}                 & 0.000                    & 0.028            \\
                        &               & Ens. Acc.           & 6          & 0.350             & 0.147                         & 0.364                       & \resub{0.650}                 & 0.187                    & 0.508            \\\midrule
Next Panel Easy         & Baseline      & Circ. Acc.          & 3          & 0.000             & 0.000                         & \resub{0.333}               & 0.281                         & 0.000                    & 0.025            \\
                        & Left-First    & Circ. Acc.          & 3          & 0.000             & 0.000                         & \resub{0.333}               & 0.234                         & 0.000                    & 0.068            \\
                        & Cropped       & Circ. Acc.          & 3          & 0.000             & 0.000                         & \resub{0.333}               & 0.263                         & 0.000                    & 0.019            \\
                        & With Text     & Circ. Acc.          & 3          & 0.003             & 0.003                         & 0.244                       & \resu{0.414}                  & 0.018                    & \resb{0.759}     \\
Next Panel Diff.        & Baseline      & Circ. Acc.          & 3          & 0.000             & 0.000                         & \resub{0.333}               & 0.275                         & 0.003                    & 0.006            \\
                        & Left-First    & Circ. Acc.          & 3          & 0.000             & 0.000                         & \resub{0.333}               & 0.213                         & 0.000                    & 0.034            \\
                        & Cropped       & Circ. Acc.          & 3          & 0.000             & 0.000                         & \resub{0.333}               & 0.225                         & 0.000                    & 0.029            \\
                        & With Text     & Circ. Acc.          & 3          & 0.000             & 0.006                         & 0.244                       & \resu{0.367}                  & 0.000                    & \resb{0.612}     \\
\bottomrule
\end{tabular}
\end{table*}
}

The entire benchmark results are summarized in Table~\ref{tab:benchmark-overview}.
Figure~\ref{fig:radarchart} presents these values as radar charts.
Overall, the Single-Panel Recognition task performance is high for all models,
despite the differences in the image style between natural images and manga.
Most models struggle with the Multi-Panel tasks.

\subsection{Character Count}
\notable{
\begin{table*}[!t]
    \fontsize{8pt}{9.5pt}\selectfont
    \caption{Per-label accuracy of the Character Count tasks.
    Boldface numbers indicate the highest score among all models.
    Underlined numbers indicate the highest score among the open-source models.
    \label{tab:character-count}}
    \vspace{8pt}
    \centering
    \begin{tabular}{lc|ccccc|c}
    \toprule

Task                & Label           & CogVLM              & \makecell{EvoVLM\\JP-v1-7B}   & \makecell{LLaVA\\1.5-13B}   & \makecell{LLaVA\\NeXT-34B} & \makecell{Qwen-VL\\Chat} & GPT-4o           \\\midrule
Character Count     & 0               & 0.425               & 0.007                         & 0.671                       & \resu{0.705}               & 0.007                    & \resb{0.938}     \\
                    & 1               & \resu{0.993}        & 0.449                         & 0.914                       & 0.983                      & 0.746                    & \resb{0.997}     \\
                    & 2               & \resub{0.993}       & 0.961                         & 0.918                       & 0.987                      & 0.918                    & \resb{0.993}     \\
                    & 3               & \resub{0.970}       & 0.439                         & 0.810                       & 0.920                      & 0.840                    & 0.966            \\
                    & 4               & 0.759               & 0.358                         & 0.642                       & \resu{0.869}               & 0.723                    & \resb{0.920}     \\\midrule
Macro Average       &                 & 0.828               & 0.443                         & 0.791                       & \resu{0.893}               & 0.647                    & \resb{0.963}     \\
\bottomrule
\end{tabular}
\end{table*}
}

Table~\ref{tab:character-count} shows a label-wise accuracy of the Character Count task.
CogVLM has a label-wise accuracy within 0.004 compared to GPT-4o's performance
on labels ``1,'' ``2,'' and ``3,''
even exceeding GPT-4o on label ``3.''
However, the open-source models have a lower accuracy for the label ``0'' which is the main sources of the performance difference compared to GPT-4o.

The prompt used for this task did not directly state that no characters might be shown (``\textit{How many manga characters are visible in this manga panel?}''), therefore the open source models might not be expecting ``0'' as a possible answer.
The CircularEval performance for the label ``0'' provides insights into each model's capability for interpreting and answering questions without a finite set of fixed answers.

\subsection{The Impact of Onomatopoeia in Scene Understanding}
For EvoVLM-JP and LLaVA-NeXT, the CircularEval accuracy metric largely improves in the With Text and Crop+Text conditions compared to the Baseline condition.
In these conditions, transcriptions of all visible onomatopoeia in the panel are included in the prompt.
This suggests that while these models cannot perform Japanese onomatopoeia OCR,
they can understand onomatopoeia when provided as text through the prompt.
This highlights the multimodal nature of our benchmark, allowing separate assessment of the visual and language modalities.

In the Onomatopoeia Scene task, GPT-4o performs
with a CircularEval metric of 0.958 for the Baseline condition and 0.826 for the Cropped condition
where the onomatopoeia is cropped out from the image.
For the open-source models, the same level of performance drops were not visible.
We suspect that this performance gap is due to the fact that in some panels,
the onomatopoeia text is necessary for deducing the correct answer.

\subsection{Prompt Sensitivity on Multi-Panel Tasks}
GPT-4o does not achieve a high score on the CircularEval metric for the Next Panel Inference task.
To describe this performance, we conduct a prompt sensitivity analysis for GPT-4o on this task.
We first gather a subset of prompts in the Baseline-Easy condition where the choices are
``A. Bottom left,''
``B. Bottom middle,'' and
``C. Bottom right,''
presented in this order.
The accuracy for this subset was 0.906.
However, when we gather all prompts
that have ``A. Bottom middle,''
``B. Bottom right,'' and
``C. Bottom left'' as the choices,
the accuracy for this subset drops to 0.070.
Among all possible permutations, these were the choices that attained the highest and lowest accuracies.
A similar analysis for the Baseline-Difficult condition showed that the accuracy ranges
between 0.694 and 0.125,
with the maximum and minimum attained by the exact same prompts.
These results suggest that GPT-4o has a strong sensitivity against the choice permutations for this task.

Meanwhile, GPT-4o scores above the task's chance rate in CircularEval for the Emotion task,
despite the fact that it has more choices than the Next Panel Inference task.
One key difference between these tasks is that the choices in the latter task have an inherent order since they consist of positional keywords.
We hypothesize that the presence of this inherent order contributes to GPT-4o's prompt sensitivity.

This is also supported by GPT-4o's low CircularEval score on the Panel Localization task, which also have positional choices.
Although an analysis over all permutations is infeasible for this task due to the high number of combinations,
the fact that Ensemble Accuracy is higher than the chance rate for this task suggests that GPT-4o struggles for certain permutations in this task as well.

This performace drop effect mostly vanishes for the ``With Text'' condition.
Here, the positional keywords are added in the prompt when providing information about the dialogues contained in each choice panel.
This information is not included in the Baseline prompt.
We suspect that the reduced performance drop is due to this explicit appearance of the positional keywords,
making it easier for GPT-4o to relate the choices directly to the dialogues without reasoning about the ordering of the choices.

\subsection{Limitations of CircularEval}
For the Next Panel Inference task, LLaVA-1.5 achieves a CircularEval score of 0.333 in all conditions except With Text.
This is because in these experiments, LLaVA-1.5 always answered ``Bottom middle'' for all choice permutations,
leading to a perfect score for questions with ``Bottom middle'' as the answer.
Although this score is higher than all other models, it clearly shows that the model is not performing well at the task,
highlighting a limitation of CircularEval when assessing tasks that are too difficult.

CircularEval correctly penalizes models that always answer a constant index.
For example, in the Next Panel Inference Easy Baseline condition,
CogVLM always answered ``B,'' and EvoVLM-JP never answered ``A,''
resulting in a CircularEval score of zero.

\subsection{Story Comprehension Capability}
In the With Text condition of the Next Panel Inference task, where dialogue is provided in the prompt text,
LLaVA-NeXT is the only open-source model to score above the chance rate of 0.333 in both Easy and Difficult settings.
This indicates that while LLaVA-NeXT struggles to recognize the Japanese dialogue text within the image,
it has some capability to comprehend and reason about the story when given the dialogue as text throught the prompt.

\subsection{Visual Content Comparison Capability}
In the Easy setting for the Next Panel Inference task, wrong answers are sampled from different manga,
creating many visual differences between correct and wrong answers.
This offers the model substantial room to rely on visual cues.
However, in the Baseline setting, models could not score above the chance rate for this task,
even for models that performed above the chance rate in the Panel Localization task.

We hypothesize that this is due to the models' capabilities of comparing contents within an image.
The Panel Localization task requires choosing a panel with a given label, allowing room for it to be solved by inspecting each panel individually.
However, for the Next Panel Inference task, the model must compare the contents between panels to utilize the visual cues.
We suspect that this added complexity may be a major challenge for open-source models, indicating significant room for improvement in LMMs for manga understanding tasks.
As discussed in the Benchmark Design Overview, comprehending two-dimensional panel layouts is crucial for understanding the visual language of manga, highlighting this as an important area for improvement in LMMs.

\section{Conclusion}
In this paper, we proposed MangaUB, a manga understanding benchmark for LMMs that allows a fine-grained analysis for the various capabilities required for manga understanding tasks.
The results show that while recognition tasks for single panels have a high capability of recognizing the content of single panels,
the understanding of emotions and two-dimensional panel layouts has a large margin for improvement.
Our analysis suggests that current open-source LMMs struggle with comparing the content of multiple panels shown in a single picture.
The Onomatopoeia Scene task shows that while the open-source models that we have experimented in this paper struggles with performing Japanese onomatopoeia (sound effect) OCR, some of the models are capable of reading Japanese itself, when the text is given through the prompt text.
This shows the benchmark's capability to individually assess the performance of both the visual and language modalities.
We believe our benchmark will become a valuable tool for advancing mulitmodal processing of manga.

\section*{Acknowledgments}
We thank Yusuke Matsui, Jeonghun Baek, Yingxuan Li, Atsuyuki Miyai,
Naohiro Yanase, and Yu Kashima for the fruitful discussions for this paper.
Parts of the helper functions in the benchmark code were generated by GPT-4o, an AI system published by OpenAI.

\begin{Biography}{Hikaru Ikuta}{\,} is pursuing his Ph.D in Information Science and Technology at The University of Tokyo, Tokyo, Japan.
Contact him at ikuta@hal.t.u-tokyo.ac.jp.
\end{Biography}

\begin{Biography}{Leslie Wöhler}{\,}
is a postdoctoral researcher at the Dept. of Information and Communication Eng., The University of Tokyo, Tokyo, Japan.
Contact her at woehler@hal.t.u-tokyo.ac.jp.
\end{Biography}

\begin{Biography}{Kiyoharu Aizawa}{\,} (S'83 -- M'88 -- F'16) is a Professor with the Dept. of Information and Communication Eng., The University of Tokyo, Tokyo, Japan.
Contact him at aizawa@hal.t.u-tokyo.ac.jp.
\vadjust{\vfill\pagebreak}
\end{Biography}

\end{document}